\documentclass{esannV2}
\usepackage[dvips]{graphicx}
\usepackage[latin1]{inputenc}
\usepackage{amssymb,amsmath,array}
\usepackage{algcompatible}

\newcommand{\ket}[1]{\left|{#1}\right\rangle}
\newcommand{\bra}[1]{\langle{#1}|}
\newcommand{\inner}[2]{\langle{#1}|{#2}\rangle}

\begin{document}
\title{Supervised quantum gate ``teaching'' for quantum hardware design}

\author{Leonardo Banchi$^1$, Nicola Pancotti$^2$ and Sougato Bose$^1$
%
\thanks{L.B. and S.B. acknowledge the financial support
by the ERC under Starting Grant 308253 PACOMANEDIA.}
%
\vspace{.3cm}\\
%
1- 
Department of Physics and Astronomy, 
University College London, \\
Gower Street, London WC1E 6BT, United Kingdom
%
\vspace{.1cm}\\
2- Max-Planck-Institut f\"ur Quantenoptik, \\
Hans-Kopfermann-Stra{\ss}e 1, 85748
Garching, Germany
}

\maketitle

\begin{abstract}
  We show how to train a quantum network of pairwise interacting qubits 
  such that its evolution implements a target quantum algorithm into 
  a given network subset.
  Our strategy is inspired by supervised learning and is designed to help
  the physical construction of a quantum computer which operates with 
  minimal external classical control. 
\end{abstract}

\section{Quantum networks for computation}
A quantum computer is a device which uses peculiar quantum effects,
such as superposition and entanglement, to process information
\cite{nielsen2010quantum}. 
Although current quantum computers operate on few
quantum bits ({\it qubits}) \cite{ladd2010quantum,coffey2014incremental}, 
it is known that large-scale quantum devices 
can run certain algorithms exponentially faster than the classical 
or probabilistic counterpart \cite{nielsen2010quantum}. 

Recently there has been many proposals to speed-up machine learning strategies 
using quantum devices \cite{wittek2014quantum}. Most of these proposals either
use quantum algorithms to achieve faster learning \cite{rebentrost2014quantum} 
or exploit quantum fluctuations to escape from local minima in training 
the Boltzmann machine
\cite{denil2011toward}. 
In this paper we consider a different perspective. 
Since the actual development of a general quantum computer is still in its
infancy, 
rather than focusing on the advantages of quantum devices for machine learning 
we show how machine learning can help the construction of a quantum computer. 

To keep the discussion realistic, we focus on a superconducting quantum
computing architecture \cite{mariantoni2011implementing,chen2014qubit} 
where each qubit is realized with a superconducting circuit cooled at low 
temperature. The pairwise coupling between two qubits is introduced 
by connecting them via a capacitor (or an inductor) whose strength can be
tuned by design. Given the flexibility in wiring the different qubits, it is
then possible to build a quantum network with tunable couplings.

In the next sections we introduce basic aspects of the physical simulation of 
quantum operations to set up the formalism and then we propose our strategy which is 
inspired by supervised learning.

\subsection{Physical implementation of quantum gates}
From the mathematical point of view each qubit is described by a two dimensional
Hilbert space $\mathbb{C}^2$, while the Hilbert space of $N$ qubits is given the 
the tensor product $\mathbb{C}^2\otimes\mathbb{C}^2\dots=\mathbb{C}^{2^N}\equiv
\mathcal{H}_N$. 
The possible quantum states, like the state $00101$ for a classical 5-bit register, 
correspond to vectors of unit norm in the Hilbert space.  An arbitrary operation,
namely a quantum {\it gate}, corresponds to a unitary matrix $U$ acting on
$\mathcal{H}_{N}$. In more physical terms, $U$ is the solution of the 
Schr\"odinger equation $i \frac{\partial U}{\partial t} = HU$, where $i$ is the
imaginary unit, $t$ represents time, and $H$ is the Hamiltonian, a 
Hermitian $2^N\times
2^N$ matrix which describes the physical interactions between the qubits. 
If the qubits are unmodulated, namely there is no external time-dependent
control so $H$ is independent on $t$, 
then after a certain time $t$ the operation on the qubits is 
given by 
$U=e^{-itH}$, being $e^{(\cdot)}$ the matrix exponential. 
In principle, for any given operation $U$ there are some corresponding interactions 
modeled by a Hamiltonian $H$ so that $U=e^{-itH}$. However, in 
physical implementations of quantum computers \cite{ladd2010quantum}
the range of possible Hamiltonians is severely limited, thus drastically
restricting the range of achievable quantum operations without external 
control.  
This problem is typically solved by switching on and off different interactions 
in time so the final operation is the product $U_1U_2\dots$ where $U_n=e^{-it_n
H_n}$, being  $H_n$ a sequence of interaction Hamiltonians, each one 
switched on for a time $t_n$. 
As in classical computation, there is a minimal set of gates $\{U_n\}$ which
enable {\it universal} quantum computation simply by concatenating at different
times gates from this
set \cite{barenco1995elementary}. Indeed, most quantum algorithms are nothing
but a known sequence of universal operations. 
The implementation of this sequence however 
requires an outstanding experimental ability
to perfectly switch on and off different physical couplings at given times. 
Possible errors or 
imperfections in this sequential process accumulate in time and may affect the outcome, if 
not tacked with error correcting codes \cite{nielsen2010quantum}. 

Given this experimental difficulty, it is worth asking whether the unitary $U$ 
which results from a recurring 
sub-sequence of the algorithm can be implemented directly in
{\it hardware}. 
Indeed in
 the next section we propose a different strategy which exploits auxiliary 
qubits to implement quantum operations with physical interactions and no
external control. 
This strategy could potentially allow an experimentalist to create a quantum
device which, by  
simply ``waiting'' for the natural dynamics of the network, 
is able to implement transformations like the Quantum Fourier Transform
which are ubiquitous in quantum algorithms, and may even provide an alternative
paradigm for general quantum computation. 
Our method shares similar goals with a recent proposal by Childs
\cite{childs2013universal}, but it 
is completely different because it uses weighted networks which allow 
us to significantly lower the number ancillary qubits required for the
operation.

\subsection{Engineered unmodulated networks for computation}
Our aim is to implement a quantum operation (a unitary%
\footnote{We consider a quantum gate, 
but our formalism can be easily extended to more general quantum channels
\cite{nielsen2010quantum}.} 
matrix $U$) on a
$N$ qubit register exploiting the physical interactions available for that
hardware architecture, and avoiding to use external control fields. 
Since a generic $2^N\times 2^N$ gate $U$ is impossible to realize 
with $N$ qubits and pairwise interactions only, 
as described before, we consider a larger network of
$N'>N$ qubits and we engineer the strength of the pairwise interactions between
them to implement the operation, when possible. 

This problem shares some similarities with supervised learning  in artificial neural
networks although, as we clarify in the following, is also very different in some aspects. 
In supervised learning, given a training set $\{I_k, O_k\}_{k=1,\dots,M}$ the goal is
the find a functional approximation $O_k=f(I_k)$ which is also able to predict the
output corresponding to unknown inputs missing from the training set. 
Even when there is no prior knowledge of $f$, it is possible to approximate the 
input/output relations with a neural network composed by input, output and hidden
layers \cite{bishop2006pattern}. The learning procedure then consists in finding 
the optimal weights between nodes of different layers such that the desired
input/output relation is reconstructed. 

\begin{figure}[h!]
  \centering
  \includegraphics[width=0.7\textwidth]{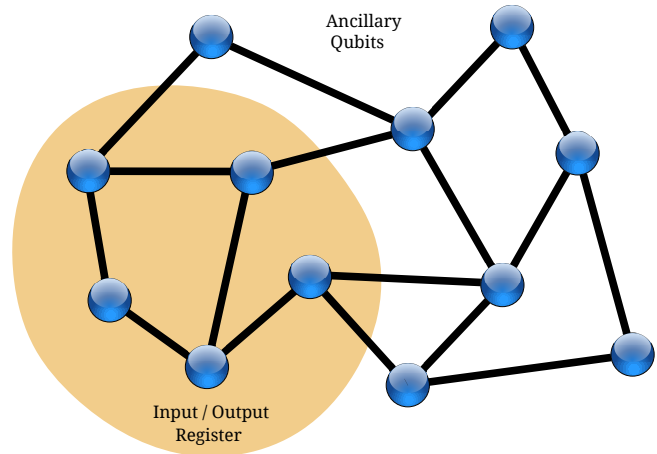}
  \caption{Example of a quantum network composed of register and ancillary qubits. Each
  vertex is a qubit and each edge represents pairwise interactions. The target
  operation $U$ is implemented only on the register qubits. }
  \label{fig:qnet}
\end{figure}

On the other hand, 
in our problem the functional relation between inputs and outputs, namely the
gate $U$ is already known in advance. However, the corresponding Hamiltonian 
may contain simultaneous interactions between $3$ or more qubits which 
unlikely appear in physical implementations. To simulate $U$ using pairwise
interactions only, we consider then a larger network as in Fig.~\ref{fig:qnet}
with ancillary qubits playing the role of the hidden layers,
and we train the weights, namely the physical
couplings, such that the dynamics of the network reproduces $U$ in a given
subset of qubits. To simplify the implementation we
assume that the input and output registers are made of the same physical qubits,
although this assumption may be removed. 
In the next section we show how to formally model the training procedure.

\section{Supervised gate ``teaching'' }
In supervised learning the training set is composed of data, whose input to
output map $f$ is not known. On the other hand, we know the gate $U$ and we want 
find find the physical couplings, i.e. the weights of the network in Fig.~
\ref{fig:qnet}, such that the quantum network evolution implements $U$ in the
register. Because of this difference we named our strategy 
``teaching'' rather than learning. 
In our case we can build an arbitrary large training set by choosing random 
input states\footnote{
In the ``bra and ket'' notation
\cite{nielsen2010quantum} $\ket{\psi}$ refers to a normalized 
vector.}
$\ket{\psi_j}\in\mathcal{H}_N$ 
and finding the  corresponding output states
$\mathcal{H}_N\ni\ket{\psi_j'}=U\ket{\psi_j}$, so the 
generated $M$-dimensional ($M$ being variable) training set is 
\begin{equation}
  \mathcal T = \{ (\ket{\psi_j}, U\ket{\psi_j}) \quad : \quad j=1,\dots,M\}~.
  \label{training}
\end{equation}
In principle the optimal inputs may depend on the target gate $U$. 
However, for simplicity and generality, in \eqref{training} 
each $\ket{\psi_i}$ is sampled from the Haar measure 
\cite{banchi2015quantum}. 

The quality of the implementation of the target gate $U$ into the dynamics of 
the quantum network can be measured by defining a {\it cost} function which, for any
input state $\ket{\psi_j}\in\mathcal T$, measures the distance between the
output state of the evolution and the expected output $U\ket{\psi_j}$. 
The similarity between two quantum states $\ket{\psi}$ and $\ket{\phi}$ 
is measured by the fidelity%
\footnote{The ``bra'' $\bra{\psi}$ corresponds to the Hermitian conjugate 
$\ket{\psi}^\dagger$ so 
$\inner{\psi}{\phi}$ is the inner product between the vectors $\ket{\psi}$ and
$\ket{\phi}$. 
}
$|\inner{\psi}{\phi}|^2$. 
Because of the Cauchy-Schwarz inequality, it is 
$0\le|\inner{\psi}{\phi}|^2\le1$ and the upper value 
$|\inner{\psi}{\phi}|=1$ is obtained only when $\ket\psi=\ket\phi$. 
However, because of entanglement between the register and the ancillary qubits 
the state of the register is not exactly known, but it can be in different 
states $\ket{\phi_j}$ with probability $p_j$. Such a {\it mixed} state is
mathematically described by the matrix \cite{nielsen2010quantum}
$\rho=\sum_j p_j \ket{\phi_j}\bra{\phi_j}$  and the fidelity between $\rho$ and 
$\psi$ becomes $\bra{\psi}\rho\ket{\psi}\equiv\sum_j
p_j|\inner{\psi}{\phi_j}|^2$. 

Given an initial state $\ket{\psi}$ let us call $\mathcal E_w[{\psi}]$ the 
state of the register after the evolution 
generated by the interaction
Hamiltonian $H(w)$ which depends on the weights $w$. This state can be
constructed by 
(i) initializing the $N$ register qubits in the state $\ket\psi$; 
(ii) setting the remaining $N'-N$ ancillary qubits in a (fixed) state
$\ket\alpha$; 
(iii) switching on the evolution described by $H(w)$ for a certain time $t$ --
without loss of generality we set $t=1$, since $e^{-itH(w)}= e^{-iH(tw)}$; 
(iv) observing the state of the register%
\footnote{Formally, steps (i) and (ii) correspond to the preparation of the 
  state $\ket{\eta_0}=\ket{\psi}\otimes\ket{\alpha}$. Step (iii) gives the state
  $\ket{\eta_t}=e^{-itH(w)}\ket{\eta_0}$. While step (iv) produces a mixed 
  state since the dynamics of a reduced system is obtained with the 
  {\it partial trace} \cite{nielsen2010quantum}, so 
  $\mathcal E_w[\psi] = {\rm Tr}_{\rm ancilla} \ket{\eta_t}\bra{\eta_t}$. 
}.
The goal is then to find the optimal weights $w$ (if they exist) 
such that the state $\mathcal E_w[\psi_j]$ is equal to the target 
state $U\ket{\psi_j}$ for each $\ket{\psi_j}\in\mathcal T$. Mathematically this
corresponds to the maximization of the average fidelity 
\begin{equation}
  F(w) = \frac{1}{M}\sum_{\ket{\psi_j}\in \mathcal T} \bra{\psi_j}U^\dagger\mathcal E_w[\psi_j]
  U\ket{\psi_j}~.
  \label{fidelity}
\end{equation}
The function $F(w)$ measures on average how the dynamics of the network 
reproduces in the register the target quantum gate $U$. 
It is non-convex in general and may have many local maxima with $F<1$.
However, an optimal configuration $\tilde w$ 
is interesting for practical purposes only if 
the error $\epsilon=1-F(\tilde w)$ is smaller than the 
desired threshold \cite{Martinis2015} (say $10^{-3}$ -- $10^{-4}$), so in
the following we say that a solution exists if an optimal $\tilde w$ is found
with $\epsilon<10^{-3}$. There are no known theoretical tools to establish 
in advance whether this high-fidelity solution $\tilde w$ can exist, 
so one has to rely on numerical methods. 
In the next section we discuss a simple algorithm for finding $\tilde w$.

\subsection{A simple algorithm for network optimization}
The explicit form of the fidelity function ($1-F$ is a cost function) as a sum
over the training set allows us to use the stochastic gradient descent training
algorithm, which is widely used for in training artificial neural networks 
within the backpropagation algorithm. However, since our training set is
variable and can be sampled from the Haar distribution of pure states we propose
an adapted version of the stochastic gradient descent with online training: 
\begin{algorithmic}[1]
  \State Choose the initial weights $w$ (e.g. at random);
  \State choose an initial learning rate $\kappa$;
  \REPEAT
  \State generate a random $\ket\psi$ from the Haar measure;
  \FOR{$j=1,\dots,L$}
  \State update the weights as
  \begin{align}
    w\to w  + \kappa\nabla_{w}\bra\psi U^\dagger
    \mathcal E_w \left[\psi\right]U\ket\psi;
    \label{e.grad}
  \end{align}
  \ENDFOR
  \State decrease $\kappa$ (see below);
  \UNTIL{convergence (or maximum number of operations). }
\end{algorithmic}
In the above algorithm, the weights are updated $L$ times before changing the
state. The parameter $L$ defines the amount of deterministic steps in 
the learning procedure and
it can be set to the minimum value $1$, so that
after each iteration the state is changed, or to higher values. On the other
hand, the learning rate $\kappa$ has to decrease \cite{spall2005introduction} 
in an optimal way to assure convergence, a common choice being 
$\kappa\propto s^{-1/2}$ where $s$ is the step counter. 
On physical grounds, as we discussed extensively in Ref.
\cite{banchi2015quantum}, the stochastic fluctuations given by choosing random
quantum states at different steps enable the training procedure 
to escape from local maxima when the weights
are far from the optimal point $\tilde w$. 

Using the above algorithm, in Ref.\cite{banchi2015quantum} we considered
pairwise interactions described by physically reasonable Hamiltonians and we 
found different quantum network configurations which implement different quantum
operations, such as the quantum analogue of Toffoli and Fredkin gate. 

\section{Concluding remarks}
We are proposing an alternative strategy to the physical implementation of
quantum operations which avoids time modulation and sophisticated control pulses.
This strategy consists in enlarging the number of qubits and engineering the 
unmodulated pairwise interactions between them so that the desired operation 
is implemented into the register subset (see Fig.~\ref{fig:qnet}) 
by the natural physical evolution. Inspired by the analogy with the training of
artificial neural networks, we then propose a simple algorithm that is
suitable for finding few-qubits networks which implement some important quantum
gates. In the long term, by finding more efficient training algorithms suitable
for larger spaces, our strategy could potentially provide an alternative paradigm for 
computation where some quantum algorithms and/or many-qubit gates 
are obtained by simply ``waiting'' for
the natural dynamics of a suitably designed network.

\begin{footnotesize}


\end{footnotesize}



\begin{thebibliography}{10}

\bibitem{nielsen2010quantum}
M.~A. Nielsen and I.~L. Chuang.
\newblock {\em Quantum computation and quantum information}.
\newblock Cambridge University Press, 2000.

\bibitem{ladd2010quantum}
T.~D. Ladd et~al.
\newblock Quantum computers.
\newblock {\em Nature}, 464(7285):45--53, 2010.

\bibitem{coffey2014incremental}
V.~C. Coffey.
\newblock The incremental quest for quantum computing.
\newblock {\em Photonics Spectra}, 48(6):36--41, 2014.

\bibitem{wittek2014quantum}
P.~Wittek.
\newblock {\em Quantum machine learning: what quantum computing means to data
  mining}.
\newblock Elsevier, Oxford, 2014.

\bibitem{rebentrost2014quantum}
P.~Rebentrost, M.~Mohseni, and S.~Lloyd.
\newblock Quantum support vector machine for big data classification.
\newblock {\em Phys. Rev. Lett.}, 113(13):130503, 2014.

\bibitem{denil2011toward}
M.~Denil and N.~De~Freitas.
\newblock Toward the implementation of a quantum {RBM}.
\newblock In {\em NIPS Deep Learning and Unsupervised Feature Learning
  Workshop}, 2011.

\bibitem{mariantoni2011implementing}
M.~Mariantoni et~al.
\newblock Implementing the quantum von {N}eumann architecture with
  superconducting circuits.
\newblock {\em Science}, 334(6052):61--65, 2011.

\bibitem{chen2014qubit}
Y.~Chen et~al.
\newblock Qubit architecture with high coherence and fast tunable coupling.
\newblock {\em Phys. Rev. Lett.}, 113(22):220502, 2014.

\bibitem{barenco1995elementary}
A.~Barenco et~al.
\newblock Elementary gates for quantum computation.
\newblock {\em Phys. Rev. A}, 52(5):3457, 1995.

\bibitem{childs2013universal}
A.~M. Childs, D.~Gosset, and Z.~Webb.
\newblock Universal computation by multiparticle quantum walk.
\newblock {\em Science}, 339(6121):791--794, 2013.

\bibitem{bishop2006pattern}
C.~M. Bishop.
\newblock {\em Pattern recognition and machine learning}.
\newblock Springer, 2006.

\bibitem{banchi2015quantum}
  L. Banchi, N. Pancotti, S. Bose.
\newblock Quantum gate learning in qubit networks: Toffoli gate without
  time-dependent control.
\newblock {\em Npj Quantum Information}, 2:16019, Jul 2016.

\bibitem{Martinis2015}
J.~M. Martinis.
\newblock Qubit metrology for building a fault-tolerant quantum computer.
\newblock {\em Npj Quantum Information}, 1:15005, 2015.

\bibitem{spall2005introduction}
J.~C. Spall.
\newblock {\em Introduction to stochastic search and optimization: estimation,
  simulation, and control}, volume~65.
\newblock John Wiley \& Sons, 2005.

\end{thebibliography}
\end{document}